\documentclass[10pt,twocolumn,letterpaper]{article}

\usepackage{wacv}              

\usepackage{graphicx}
\usepackage{amsmath}
\usepackage{amssymb}
\usepackage{booktabs}
\usepackage{multirow}
\usepackage{comment}
\usepackage{float}
\usepackage{balance}
\usepackage{soul}

\usepackage[pagebackref,breaklinks,colorlinks]{hyperref}

\usepackage[capitalize]{cleveref}
\crefname{section}{Sec.}{Secs.}
\Crefname{section}{Section}{Sections}
\Crefname{table}{Table}{Tables}
\crefname{table}{Tab.}{Tabs.}

\def\wacvPaperID{3} 
\def\confName{WACV}
\def\confYear{2024}

\usepackage[accsupp]{axessibility}

\begin{document}

\title{Deepfake detection by exploiting surface anomalies: the SurFake approach}

\author{Andrea Ciamarra$^{1,4}$, Roberto Caldelli$^{3,4}$, Federico Becattini$^2$, Lorenzo Seidenari$^1$, Alberto Del Bimbo$^1$\\
\\
$^1$University of Florence, Florence, Italy, $^2$University of Siena, Siena, Italy,\\
$^3$CNIT, Florence, Italy,
$^4$Universitas Mercatorum, Rome, Italy\\
{\tt\small \{andrea.ciamarra, lorenzo.seidenari, alberto.delbimbo\}@unifi.it,}\\ 
{\tt\small federico.becattini@unisi.it, roberto.caldelli@cnit.it}
}
\maketitle

\begin{abstract}
   The ever-increasing use of synthetically generated content in different sectors of our everyday life, one for all media information, poses a strong need for deepfake detection tools in order to avoid the proliferation of altered messages. The process to identify manipulated content, in particular images and videos, is basically performed by looking for the presence of some inconsistencies and/or anomalies specifically due to the fake generation process. Different techniques exist in the scientific literature that exploit diverse ad-hoc features in order to highlight possible modifications. In this paper, we propose to investigate how deepfake creation can impact on the characteristics that the whole scene had at the time of the acquisition.
In particular, when an image (video) is captured the overall geometry of the scene (e.g. surfaces) and the acquisition process (e.g. illumination) determine a univocal environment that is directly represented by the image pixel values; all these intrinsic relations are possibly changed by the deepfake generation process. By resorting to the analysis of the characteristics of the surfaces depicted in the image it is possible to obtain a descriptor usable to train a CNN for deepfake detection: we refer to such an approach as SurFake. 
Experimental results carried out on the FF++ dataset for different kinds of deepfake forgeries and diverse deep learning models confirm that such a feature can be adopted to discriminate between pristine and altered images; furthermore, experiments witness that it can also be combined with visual data to provide a certain improvement in terms of detection accuracy.
\end{abstract}


\begin{figure}[t]
\centering
\begin{minipage}{\columnwidth}
  $\vcenter{\hbox to 0.1em{Real}}$
  \hspace*{.3in} 
  $\vcenter{\hbox{\includegraphics[width=0.89\columnwidth]{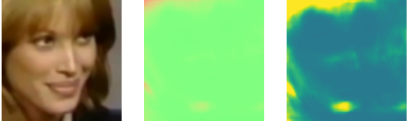}}}$ \\
  $\vcenter{\hbox to 0.1em{Fake}}$
  \hspace*{.3in}
  $\vcenter{\hbox{\includegraphics[width=0.89\columnwidth]{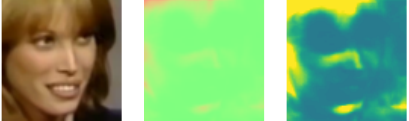}}}$
\end{minipage}
\caption{An example of surface anomalies found in fake images. From left to right: the RGB (Red-Green-Blue) face, our proposed GSD (Global Surface Descriptor) feature and the logarithm of the GSD, used here for sake of visualization to highlight the artifacts introduced by the manipulation.}
\label{fig:real_fake}
\end{figure}

\section{Introduction}
\label{sec:intro}

With the increasing of false information spreading all over the media, nowadays, trust in digital content is potentially compromised by the easiness of creating fabricated facts. Information can further undergo multiple modifications before reaching a potential user. The latest advancements of AI, especially for image and video manipulation, are fostering more and more the possibility to easily change the meaning of the information to convey, due to the fact that media content is likely to be exposed to variations of different nature. Among the possible media manipulation approaches, Deepfakes are a very recent class of methods that can generate synthetic human images. Despite having been used with astonishing results for movie production in Hollywood, Deepfakes can also be easily used for malicious purposes, such as crafting highly realistic fake propaganda.
Deepfake creation typically involves the use of deep learning to recreate some person imagery. Specifically deep networks learn how to transfer or reenact facial expression as well as how to generate a proper imitation of a person's voice and inflection.  
Nowadays, several techniques can create real-looking content easily, by using deep generative models, such as GAN-style architectures \cite{goodfellow2014generative, choi2018stargan, abdal2019image2stylegan} and diffusion probabilistic models (DPMs) \cite{ho2020denoising}.
Deepfake can be applied to different types of media, from image to video, but also fake audio can be created, e.g. through text-to-speech~\cite{kietzmann2020deepfakes}, by typing a new text, or by voice swapping~\cite{oord2016wavenet}.
Deepfake for image and video is mainly done by tampering with parts of the scene, for instance, the faces of subjects present in the media. Various techniques have been designed to alter faces, some of them regard video applications, e.g. lip syncing ~\cite{suwajanakorn2017synthesizing, prajwal2020lip} where the audio is employed to reconstruct the mouth movements over the video frames. Other general-purpose face manipulation techniques are about the visual alteration of the content, by changing the expressions or moving the face from a source image to the target one.

In this paper, we focus on detecting face manipulations in images. Face manipulations~\cite{zhang2022deepfake} can be basically summarized in two different categories: reenactment and swapping. Facial reenactment deals with manipulating certain facial attributes or reenact faces with deep learning methods while maintaining the identity unchanged. 
Face swapping~\cite{chen2019face}, instead, aims at replacing the face of a person in the reference image with the same facial shape and features of another target subject by realistically changing the identity. 
%
%
%
\begin{figure*}[ht!]
    \includegraphics[width=\textwidth]{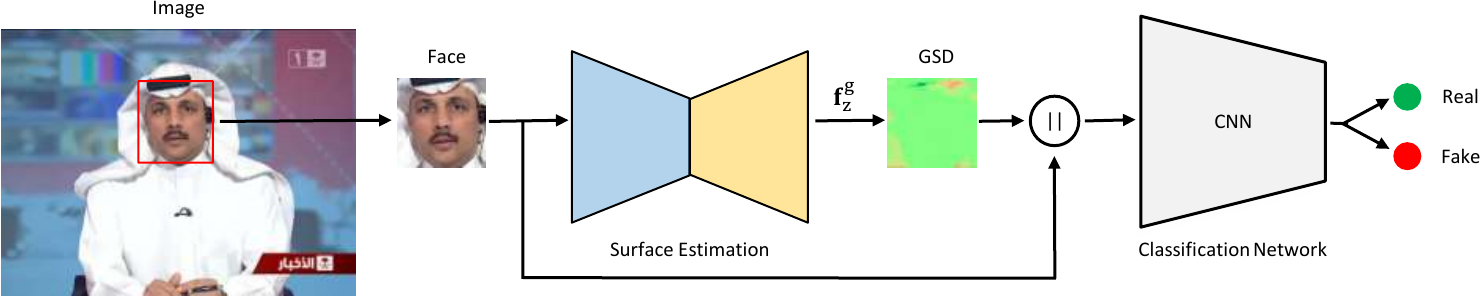}
    \caption{Pipeline of SurFake for deepfake detection. After extracting the face crop from the image, we generate its Global Surface Descriptor (GSD) through UpRightNet \cite{xian2019uprightnet} and we scale the generated vector values in $[0,255]$ to obtain an RGB image. 
    Then, we concatenate the face crop and the GSD feature at the last channel and we pass it in input to a classifier. Finally, we train the classifier to distinguish whether the content is real or fake.}
\label{fig:surfake_scheme}
\end{figure*}
According to this, it is straightforward to understand that deepfake detection is an urgent task, which prevents disinformation and avoids the diffusion of media showing people saying or doing things they never said or did.
The fundamental idea behind deepfake detection consists in the fact that a neural network during the process of generating a fake content should leave a sort of trace that is embedded as a fingerprint over the manipulated image (video). Most of the existing approaches for deepfake detection try to recover this hidden pattern to reveal false contents and to do that they generally resort to the analysis of frames at pixel level (RGB raw data).  
However, not only RGB-level inconsistencies can be detected as anomalies in the visual space, in fact many fine details can be affected by the forgery without compromising the visual perception of the image itself. More precisely, the camera acquisition process is itself a signature that is incorporated into the image. We argue that such information about the acquisition moment could be altered by deepfakes and this could be exploited for the detection task.
Such information relates to an ensemble of specific characteristics of the scene, for instance, the external illumination source, including lighting, shadows and reflections, which all impact on the surfaces present in the environment and on the objects at the time of image acquisition. Other relevant details regard the face pose, but even the camera parameters may be somehow embedded into the image and strictly related to the image capturing, e.g. lens distortions and intrinsic noises.
In the literature, some works \cite{sun2021improving, liang2023facial} tackled the deepfake detection problem from different angles, by leveraging geometrical aspects of facial landmarks, in which inconsistent or highly synthetic patterns are found in generated fakes by different forgery techniques.

The proposed approach leverages on the analysis of the features of the framed scene that are determined by the overall geometry of the scene itself (e.g. surfaces) and by the original image acquisition process (e.g. illumination, camera orientation).
Different from other research works focusing on individuating fakes by detecting specific patterns in terms of depth map or face motion, we specifically deal with the surfaces present within the acquired scene by exploiting the modifications induced to the surface normals and determined by the deepfake alteration. A general visual example of this is provided in Figure \ref{fig:real_fake} where it can be appreciated how a modification, just on the mouth of the woman, can also determine some slight global variations on the other parts of the global surface descriptor (GSD) image.   
In summary, the main contributions are listed hereafter:
\begin{itemize}
\item[(i)] we propose to utilize surface geometry features of the acquired scene to highlight inconsistent patterns revealing fake images;
\item[(ii)] we study and evaluate, in which extent, such features can constitute by themselves an effective mean to discriminate between pristine and fake contents;
\item[(iii)] we conduct experiments on different kinds of forgeries and network architectures to verify that the new proposed surface-based feature can be advantageously combined with RGB frames to get an improvement in terms of accuracy performance.
\end{itemize}
The paper is organized as it follows: after this introductory section, Section \ref{sec:related} describes main related works while Section \ref{sec:method} presents the proposed method. Section \ref{sec:experiments} is dedicated to the experimental results and Section \ref{sec:conclusions} draws conclusions providing possible future works.

\section{Related Works}
\label{sec:related}
DeepFake Detection is a recent problem raised to recognise real or tampered data, also in vision tasks, which is typically addressed as a binary classification problem. Nowadays visual content is an informative media that can be manipulated, thanks to the usage of recent generative models \cite{goodfellow2014generative, choi2018stargan, abdal2019image2stylegan, ho2020denoising}. 
In particular, the possibility of manipulating human faces has found a lot of interest both for entertainment and for malicious purposes. Several manipulation techniques can be used, either replacing faces to match the one of another subject or by simply altering facial expressions. Such manipulations are obtained, e.g. via CNNs \cite{korshunova2017fast}, conditional GANs \cite{olszewski2017realistic} or based on facial landmark alignment \cite{chen2019face}. A plethora of implementations are also available \cite{masood2023deepfakes}. 
Existing methods \cite{nguyen2022deep} for deepfake detection are designed to directly process entire videos or single frames, so to discover whether faces have been manipulated. Several works exploit handcrafted features from artifacts and inconsistencies of the fake generation process. Xian et al.\cite{xuan2019generalization} proposed to preprocess images to remove low level noise cues of GAN images, and so a forensic model is forced to learn more intrinsic features. This method allows better generalization capability than previous deepfake methods\cite{yang2017recapture, bayar2016deep}. 
Amerini et al. \cite{amerini2019deepfake} leveraged the optical flow in order to look at motion discrepancies, which are found across synthetically generated frames, and finally to classify them as original or deceptive. 
Li et al. \cite{li2020face} utilized an advanced architecture named HRNet to detect the blending boundary of Deepfakes manipulated images. Guo et al. \cite{guo2021blind} introduced a CNN model named SCnet to detect Glow-based facial forgery by learning high-level features through a hierarchical convolutional block. Zhao et al. \cite{zhao2021learning} proposed to look at source feature inconsistency within the forged image with the hypothesis that a pristine image should contain the same source features across locations. Instead of learning GAN fingerprints on fakes \cite{yu2019attributing} or visual self-inconsistencies via recorded photo metadata \cite{huh2018fighting}, Maiano et al. \cite{maiano2022depthfake} exploited depth inconsistencies located inside tampered face images to detect manipulations.
Other approaches, instead, utilize recurrent networks, e.g. RNNs or LSTMs, to look at visual artifacts within single video frames or temporal inconsistency across frames. Sabir et al.\cite{sabir2019recurrent} leveraged the use of spatio-temporal features of video streams to detect deepfakes, as temporal coherence is not innate in deepfake generation. G\"{u}era et al. \cite{guera2018deepfake} extracted frame-level features with a CNN and fed them into an LSTM to create temporal sequence descriptors, which are finally trained to be classified as real or fake. Different methods have also been proposed to exploit visual or behavioral inconsistencies, hardly removable when generating fakes. Based on the fact that a person in deepfakes has a lot less frequent blinking than that in untampered videos, Li et al. \cite{li2018ictu} proposed to crop eye areas of video frames on which features are extracted and then fed into an LSTM, and so to predict the probability of eye open or close. Caldelli et al. \cite{caldelli2021optical} proposed to leverage the optical flow to account for facial motion since artificial parts of the face contain some intrinsic dissimilarities with respect to natural expressions. Becattini et al. \cite{becattini2023head} found discrepancies in face alignment by looking at the head orientation, i.e. roll, pitch and yaw. Liang et al. \cite{liang2023facial} extract geometry facial features as peculiarities around the landmark to be discriminated between pristine and manipulated regions (e.g. spatial relationship, appearance, shape). Such features are fed into a CNN-LSTM network, and, finally, a decoder learns to map low-level features to pixel-wise manipulation localizations along with a softmax classifier to detect real and fake. Sun et al. \cite{sun2021improving} exposed abnormal facial movement patterns and time discontinuities by means of precise geometric features of facial landmarks, by making a proper calibration step, which is performed through a Lucas-Kanade operation to track landmark points and merge the detection and the prediction using a Kalman filter. Differently, we do not make use of segmentation face model \cite{liang2023facial}, facial landmarks, or calibration steps \cite{sun2021improving}. We consider the Global Surface Descriptor (GSD) of the face, which is a feature describing the geometry of the face. 
However, in contrast to \cite{becattini2023head}, which leverages head pose estimation with respect to the camera, we use a description of surface orientations at a pixel level by characterizing surface normals in a global up-right reference system that is inherently obtained along with the camera orientation.



\begin{figure*}[ht!]
\centering
    \begin{tabular}{c}
         \includegraphics[width=\textwidth]{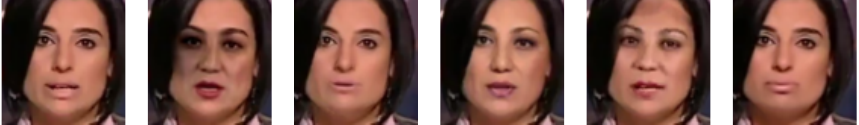} \\
         \includegraphics[width=\textwidth]{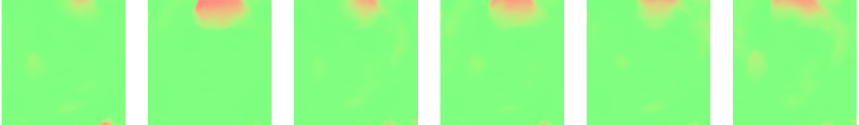} \\
         \includegraphics[width=\textwidth]{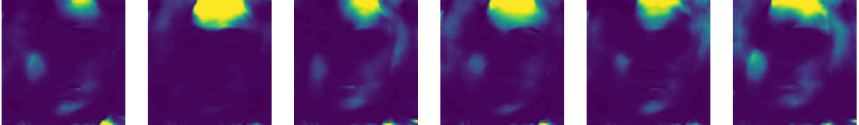} 
           \end{tabular} \\
    
    \caption{Sample frames (first row) and the corresponding Global Surface Descriptors (second row) and $\log (GSD)$ (third row) for each of the 5 different forgeries in FF++, from left to right: Real, DF, F2F, FSH, FS, NT \cite{rossler2019faceforensics++}. The third row highlights how GSD is sensitive to forgeries.}\label{fig:face_uprightnet}
\end{figure*}

\section{The proposed method}
\label{sec:method}
In this section we will introduce the proposed method, named \textit{SurFake}, which exploits inconsistencies in the features of surfaces belonging to the acquired scene to perform deepfake detection. Our pipeline is organized into three steps as depicted in Figure \ref{fig:surfake_scheme}: first, we perform face detection on each video frame using dlib \cite{king2009dlib} obtaining a face crop with a fixed resolution  of $224 \times 224$. Secondly, we run a pretrained \emph{UpRightNet}\cite{xian2019uprightnet}  on face crops to extract features (see Section \ref{sec:uprightnet_model} for details) in order to get the \emph{Global Surface Descriptor (GSD)}.
Finally, the concatenation of the RGB face crop and the GSD feature which constitutes a 6-channel tensor is used to train a deep convolutional neural network to perform the binary classification task required to detect deepfakes.

\subsection{The Global Surface Descriptor (GSD)}
\label{sec:uprightnet_model}

In order to extract subtle scene details, we face deepfake detection from a new perspective. In contrast to looking for inconsistencies in the visual perception domain, we highlight anomalies by looking into the geometrical aspects related to the camera acquisition process. Such aspects permanently mark low-level peculiarities of the image without visually affecting the content. Therefore, tampered images may contain fine-grained distortions which do not stand in the visible space. 
Typically, deepfakes depict a person in the foreground whose entire face (or some of its parts) has been tampered with. The idea behind our approach is to address forgery detection by focusing on the surface geometry of the face. To do that, we employ a deep learning model named \emph{UpRightNet} \cite{xian2019uprightnet} to estimate such geometrical characteristics presented by the oval surface of the face but also by other different surfaces such as the chin, the nose, the eye sockets or any headgear. UpRightNet is a neural network that learns to estimate the 2DoF camera orientation, i.e. roll and pitch, from a single RGB image using intermediate representations, called surface frames, estimated from both the local camera and the global up-right coordinate systems. Let us suppose to predict the per-pixel surface normals of an indoor image in the camera perspective. Surface normals on the ground and other horizontal surfaces point in the same direction as the camera up vector, instead, walls and other vertical surfaces are perpendicular to the up vector. 
Thus, camera orientation can be estimated as finding the vector which is most parallel to the ground normals and most perpendicular to the wall normals. UpRightNet solved the camera orientation problem by computing the rotation that best aligns the two estimated representations of the surface frames.
A surface geometry frame $\mathbf{F}(i)$ is estimated from each pixel $i$, as a $3 \times 3$ matrix of mutually orthogonal unit vectors, that is normals, tangents and bitangents respectively: $\mathbf{F}(i)=[\mathbf{n}(i), \mathbf{t}(i), \mathbf{b}(i)]$ with $\mathbf{n}(i), \mathbf{t}(i), \mathbf{b}(i) \in \mathbb{R}^3$. UpRightNet estimates two surface frames, one in the local camera coordinate system, $\mathbf{F}^c(i)$, and one in the global up-right coordinate system, $\mathbf{F}^g(i)$. In order to predict roll and pitch of the camera, UpRightNet aligns the up-vector in the two representations, by using the z-component of $\mathbf{F}^g(i)$, i.e.  $\mathbf{f}_z^g(i) \in \mathbb{R}^3$. 
Such alignment is computed by learning weights to solve a constrained least squared problem using ground-truth camera orientations. Due to the fact that the feature $\mathbf{f}_z^g(i)$ substantially provides a 3-channel global description of the surfaces belonging to the acquisition scene, we have considered that it could be a good candidate to possibly give evidence of a manipulation. Such a feature, denominated \emph{Global Surface Descriptor (GSD)}, will be analysed more in depth within the next sub-section.


\subsection{Analysis of Deep Geometric Representations}\label{sec:deepfake_detection}
This section presents how UpRightNet features can be useful in the context of deepfake detection;
in Figure \ref{fig:face_uprightnet} we depict an example of a pristine face crop along with various face manipulations methods implemented in FaceForensics++\cite{rossler2019faceforensics++} (first row). We also show, in the second row the corresponding GSD feature extracted by UpRightNet after passing the face image in input. Finally, in the last row, we enhance the colorization of GSD in order to visually highlight how the GSD feature could be useful to detect anomalies by depicting the logarithm of the feature. 
In fact, upon a visual inspection, the GSD features may appear similar to each other for different faces, even appearing uniform regardless of the content.
This is due to the fact that the faces are typically framed frontally in the global upright coordinate system. Similarly to global representations estimated for indoor images, the light green pixels stand for surfaces whose normals are perpendicular to the up-ward vector, e.g. the walls in a room, just like most of the face pixels too. There are also other parts of the image which are sometimes colored with shades of red and dark green, that encode surfaces parallel to the ground of a room. In our face domain, pixels with normals parallel to the ground are located at the top and at the bottom of the image. Therefore, the geometry estimated for face images is quite consistent with indoor environments. However, we would like to demonstrate how this geometry, i.e. our proposed GSD feature, is useful in our task. To this aim, we here highlight anomalies, by calculating the logarithm of each image pixel on the first of the 3 channels and we plot the results in the third row in Figure \ref{fig:face_uprightnet}, for each face manipulations.\\
First of all, we generally observe that some artifacts are added or exaggerated at the top of the image for all manipulations, which explains that any alteration produces unavoidable patterns (see the yellow color and the surrounding parts located at the top of the images in the third row, i.e. the forehead). In the face swapping approach the alterations are mainly reported around the outer facial landmarks for FS and FSH. Interestingly, the latter is also a more recent learning-based face replacing method than the well-known DF, where the face looks almost flat. As soon as facial expressions are tampered (i.e. by performing either F2F or NT) and small parts are faked, e.g. mouth or eyes, other relevant regions are also compromised, e.g. cheeks and hair. Although the GSD feature may look scarsely informative at once, we argue that such subtle details can come up more visible for a neural network trained to detect these synthetic patterns, in the sense of anomalies in the geometric estimation of the face.

\section{Experimental results}
\label{sec:experiments}
In this section, we will present the experimental results carried out in order to verify the effectiveness of the presented approach and, in particular, of the GSD feature.

\subsection{Implementation Details}
\label{subsec_res_imp}

\paragraph{Dataset}

We conduct experiments on FaceForensics++ (FF++) \cite{rossler2019faceforensics++},  one of the most widely used datasets for deepfake detection. It has collected 1000 original real videos from the internet and for each video 5 different forged versions are generated. This dataset comprises two types of face manipulation techniques: face swapping, in which the face identity in the source image is replaced with the target one, and face reenactment, in which the facial expression in the source image is altered from the one in a target image, while maintaining the identity. FaceForensics++ includes three swapping methods, DeepFakes (DF) \cite{deepfakes_github}, FaceShifter (FSH) \cite{li2020advancing} and FaceSwap (FS) \cite{faceswap}, and two reenactment methods, i.e. Face2Face (F2F) \cite{thies2016face2face} and NeuralTextures (NT) \cite{thies2019deferred}. In particular, two of these manipulations are computer graphics-based approaches (Face2Face and FaceSwap) while the other three are learning-based approaches (DeepFakes, FaceShifter and NeuralTextures). Specifically, NeuralTextures operates by only altering the mouth region, i.e. eye parts are unchanged while FaceShifter is a recent learning-based approach. It generates high fidelity identity preserving face swap results being able, differently from the other two face swapping methods, to deal with facial occlusions using a double synthesis stage. 
%

%
Overall, there are 1000 forged videos for each face manipulation, for a total of 5000. The image resolutions vary across videos, from $272 \times 480$ up to $1920 \times 1080$. 
FaceForensics++ provides raw videos, and two versions compressed using the H.264 codec, i.e. light compression (c23), which is nearly lossless, and heavy compression (c40). For all our experiments we choose the c23 videos, which is indicated from the dataset authors as HQ (high quality).

\begin{figure*}
    \begin{tabular}{ccccc}
         \includegraphics[width=0.18\linewidth]{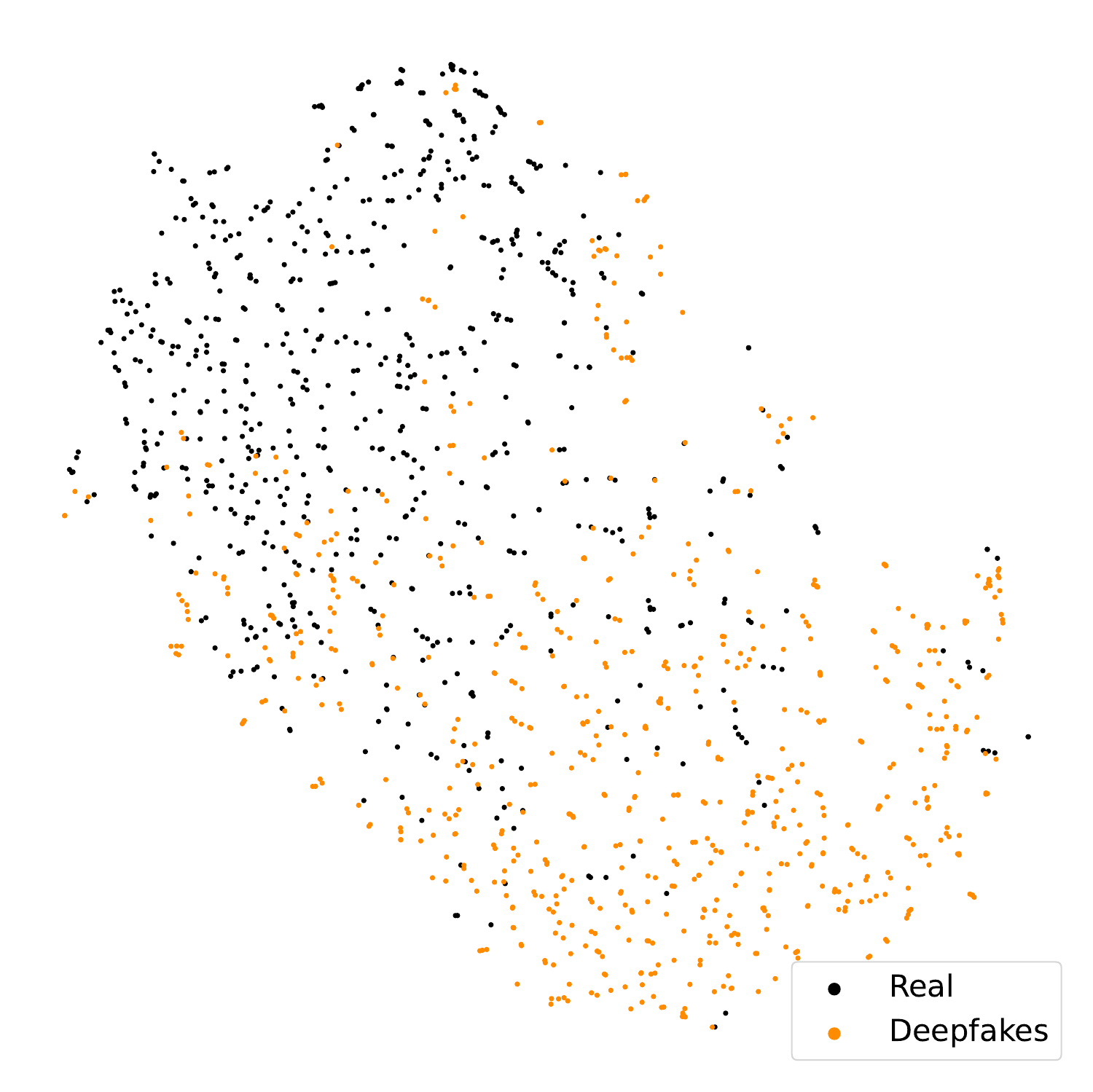} & \includegraphics[width=0.18\linewidth]{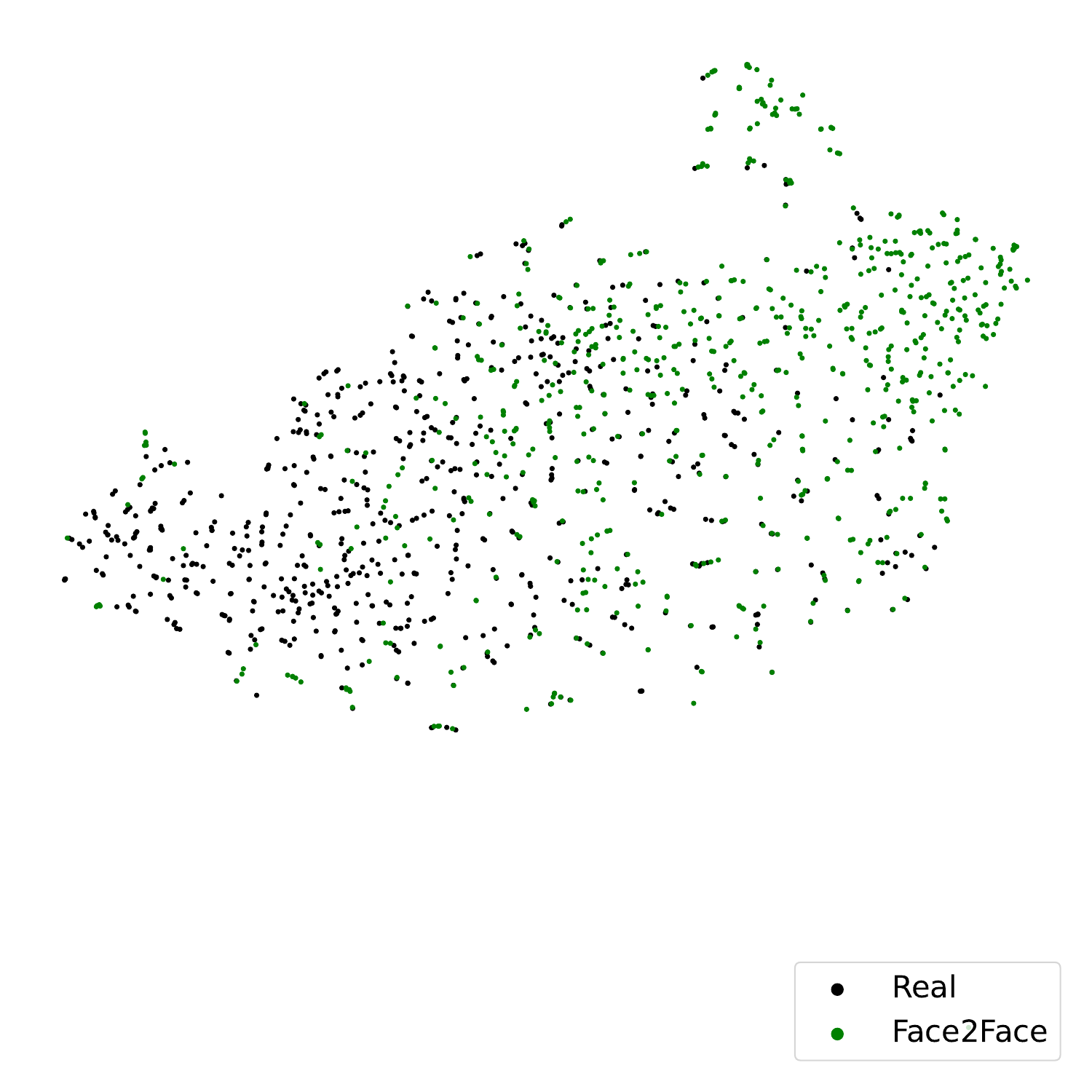} &  \includegraphics[width=0.18\linewidth]{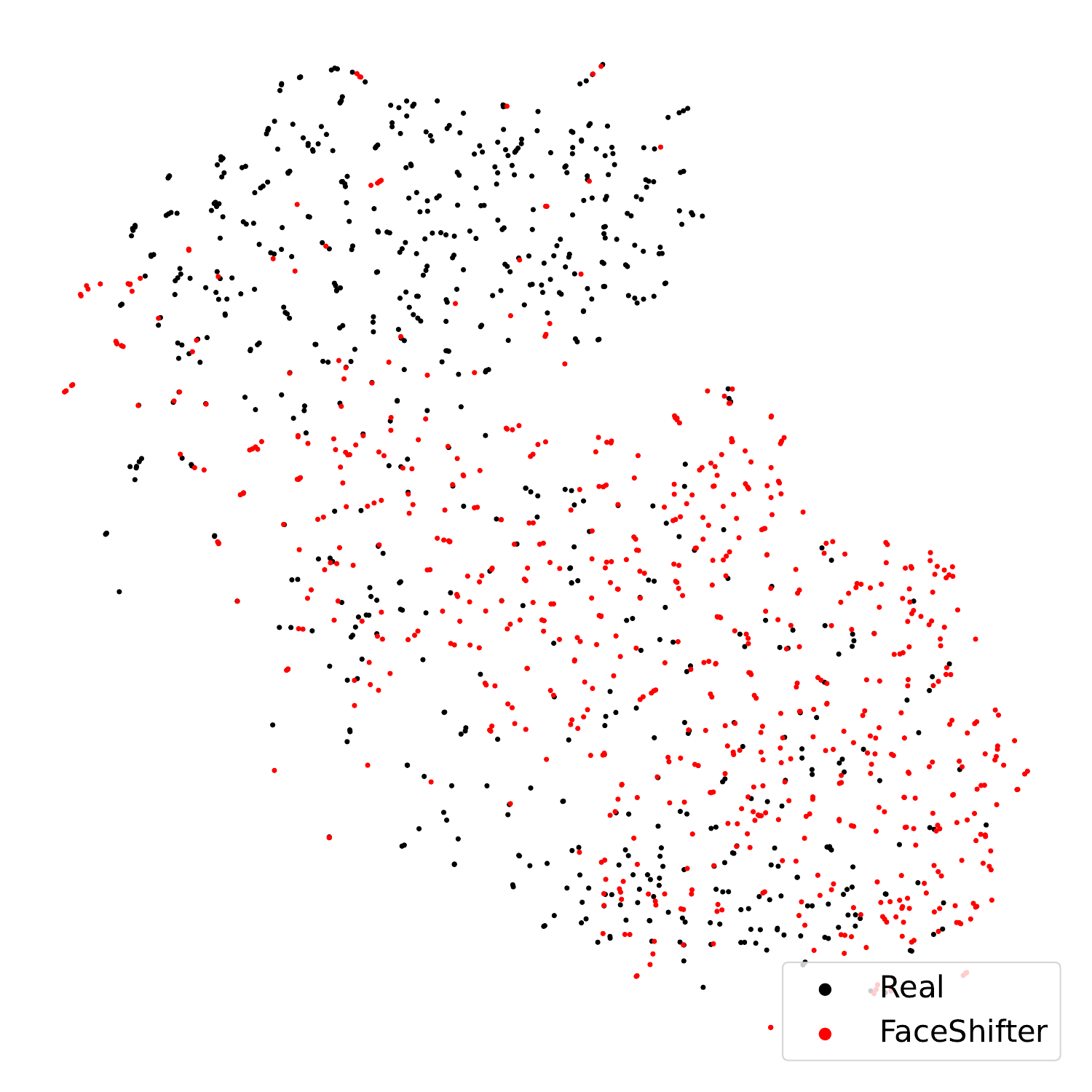} &
         \includegraphics[width=0.18\linewidth]{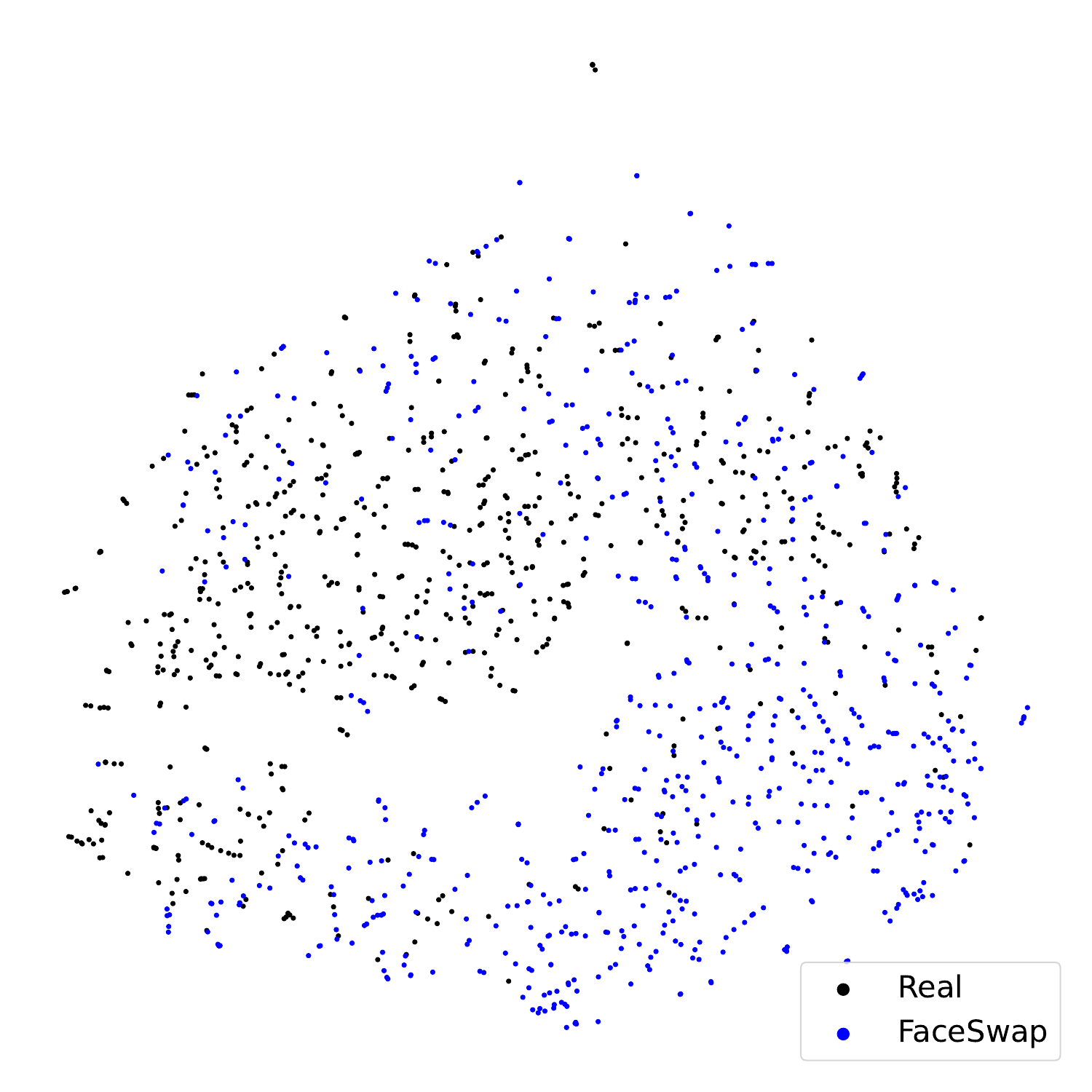} & \includegraphics[width=0.18\linewidth]{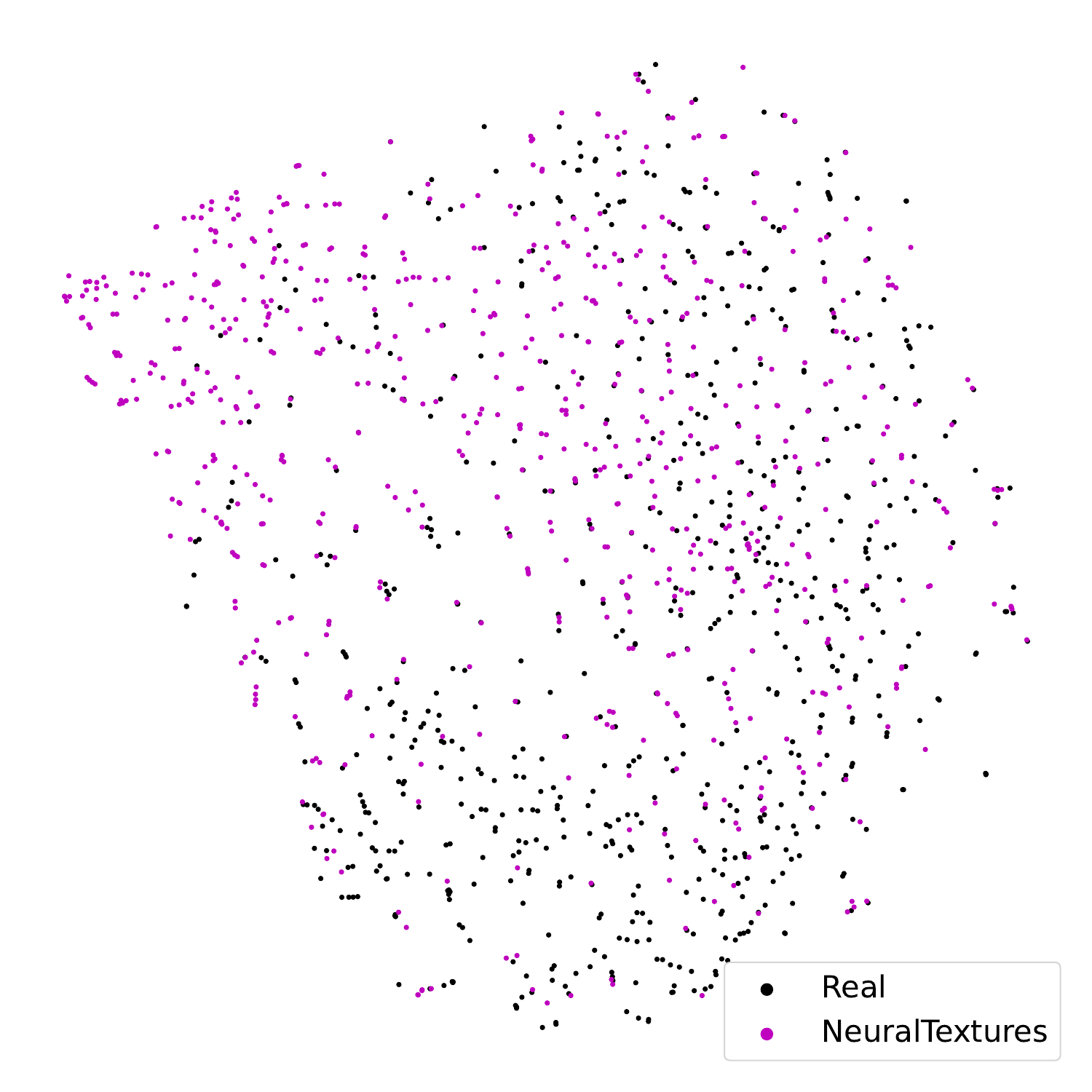} \\
    \end{tabular}
    \caption{T-SNE \cite{van2008visualizing} plots of the GSD feature activations for real and fake samples of the test set for each of the different forgeries (MobilNetV2 architecture). Only a reduced number of samples is plotted for the sake of visibility.}\label{fig:tsne_global}
\end{figure*}

\begin{figure}[t]
    \centering
    \includegraphics[width=0.8\linewidth]{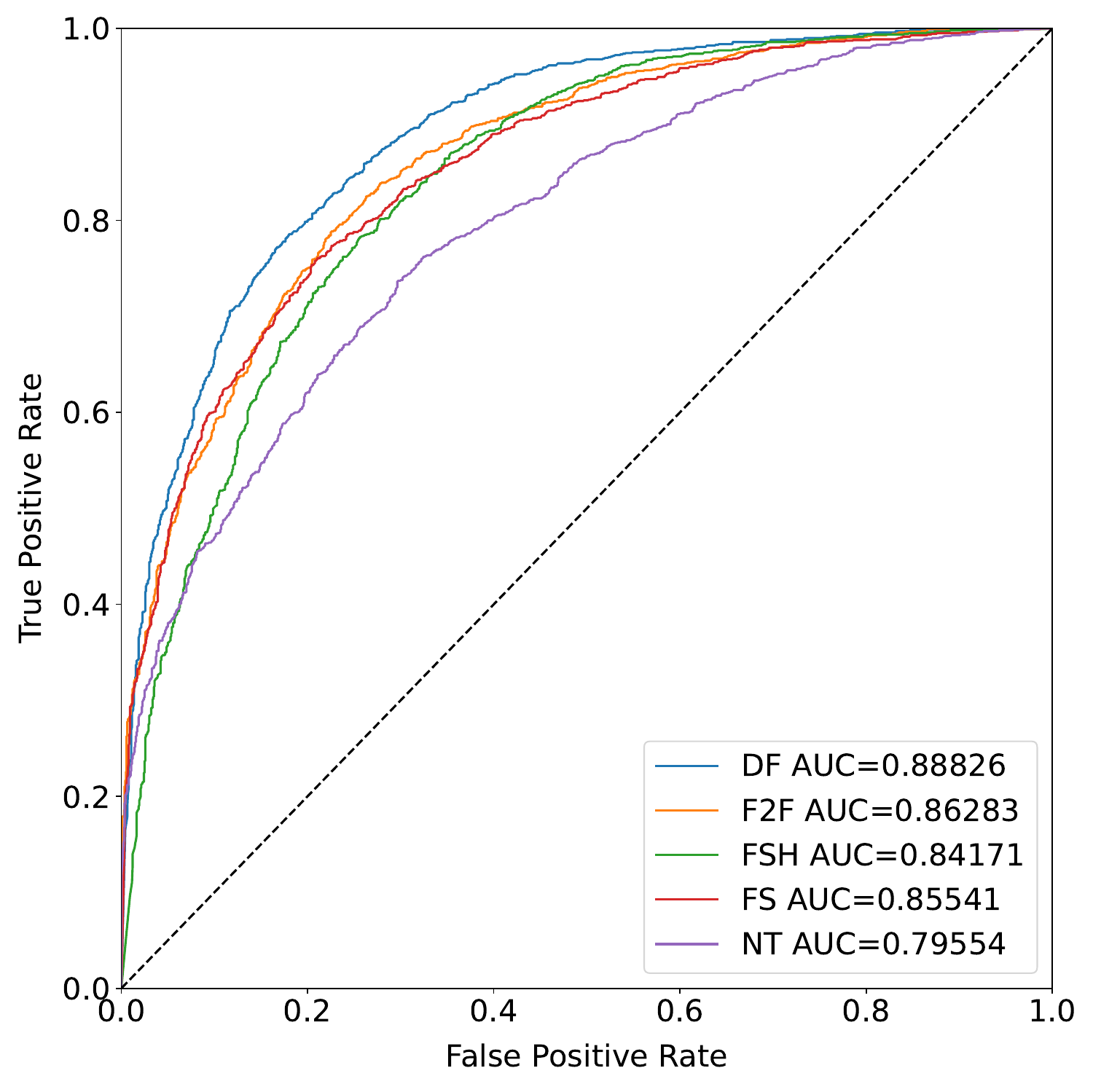} 
    \caption{ROC Curve of GSD features for real and fake using MobileNetV2 as classifier. We also reported the Area Under Curve (AUC) for each forgery.}\label{fig:roc_global_mobilenetv2}
\end{figure}

\paragraph{Data preparation}
To reduce the computational burden and exclude redundancies, we sample one frame out of ten in each video sequence.
For all our experiments we follow the 72:14:14 data split, respectively for train, validation and test sets, as indicated in \cite{rossler2019faceforensics++}, i.e. 720, 140 and 140 videos.
Following \cite{rossler2019faceforensics++}, $224 \times 224$ face crops are obtained by first extracting a $1.3$-factor enlarged crop centered at the detected face in the input image and then scaling it to the fixed resolution. We consider face crops of such a resolution as it is a standard image dimension that can be processed into most of the existing architectures. Since UpRightNet gets input images of $288 \times 384$ and generates outputs at the same resolution, we adapt face crops to such dimension in order to extract the Global Surface Descriptor (GSD) and then we rescale to $224 \times 224$, as done in \cite{xian2019uprightnet}. 
Among the UpRightNet pretrained weights on InteriorNet and ScanNet (i.e. two indoor image datasets), we chose the former one since this model estimates a more coherent representation of GSD than the latter, in accordance with the true geometry in the face domain in terms of normals perpendicular or parallel to the global upright coordinate system (see Section \ref{sec:uprightnet_model}).

\paragraph{Architectures}
In order to classify images as real or fake we train 4 different well-known and standard architectures: ResNet50\cite{he2016deep}, MobileNetV2\cite{sandler2018mobilenetv2}, EfficientNet-B0\cite{tan2019efficientnet} and Xception\cite{chollet2017xception} with pretrained weights on ImageNet.
Since we are using a neural network pretrained on classical RGB images from ImageNet, we modify the first convolutional layer, which gets the input, to handle a different type of data, i.e. the 3-band GSD feature and also a different number of input channels, as our approach is employing a total of 6, i.e. the first 3 channels for the RGB concatenated to the second 3 additional ones from GSD. Besides, to make the training more stable and to allow the model to converge faster, we make a proper weight initialization. By following \cite{maiano2022depthfake}, we calculate the average of the 
three original input channels from the pretrained model and we replace this initialization for each and every channel of GSD. We chose this initialization for all our experiments.

Because each architecture is pretrained  on ImageNet, we scale values in $[0,1]$ and then we normalize with mean $[0.485, 0.456, 0.406]$ and standard deviation $[0.229, 0.224, 0.225]$ for ResNet50, MobileNetV2 and EfficientNet-B0 respectively. For Xception, we use the Pytorch implementation and the pretrained ImageNet weights from \cite{xception_pytorch}. Since Xception accepts inputs at $299 \times 299$, we upscale our patches to fit that resolution, we scale values in $[0, 1]$ and we normalize with mean and standard deviation both set to $[0.5, 0.5, 0.5]$. Note that we apply scaling and normalization on the input values for RGB and GSD separately.

\begin{table}[t]
\scriptsize
\centering
\begin{tabular}{|c||c|c|c|c|c||c|}
\hline
& \multicolumn{6}{c|}{\textbf{FF++ forgeries}} \\ 
\hline
\textbf{Architectures} & DF & F2F & FSH & FS & NT & Avg \\ \hline \hline
ResNet50 & 0.766 & 0.725 & 0.735 & 0.690 & 0.674 & 0.718 \\
MobileNetV2 & 0.800 & 0.773 & 0.756 & 0.764 & 0.713 & 0.761 \\
EfficientNet-B0 & 0.802 & 0.773 & 0.761 & 0.754 & 0.707 & 0.759 \\ 
Xception & 0.796 & 0.759 & 0.759 & 0.747 & 0.726 & 0.757 \\ \hline 
\end{tabular}
\caption{Performance in terms of accuracy for the GSD feature on the test set with respect to the different network architectures.}
\label{tab_featacc}
\end{table}

\begin{table*}
\scriptsize
\centering
\begin{tabular}{|c||cc|cc|cc|cc|cc|cc|}
\hline
& \multicolumn{12}{c|}{\textbf{FF++ forgeries}} \\ \hline
\multirow[c]{2}{*}{\textbf{Architectures}} & \multicolumn{2}{c|}{DF} & \multicolumn{2}{c|}{F2F} & \multicolumn{2}{c|}{FSH} & \multicolumn{2}{c|}{FS} & \multicolumn{2}{c|}{NT} & \multicolumn{2}{c|}{Average} \\ \cline{2-13}
& RGB & RGB+GSD & RGB & RGB+GSD & RGB & RGB+GSD & RGB & RGB+GSD & RGB & RGB+GSD & RGB & RGB+GSD \\ \hline \hline
ResNet50 & 0.981 & \textbf{0.984} & 0.988 & \textbf{0.989} & \textbf{0.980} & 0.971 & 0.985 & \textbf{0.986} & 0.938 & \textbf{0.947} & 0.974 & \textbf{0.975} \\
MobileNetV2 & 0.987 & \textbf{0.992} & \textbf{0.990} & 0.989 & 0.985 & \textbf{0.989} & \textbf{0.990} & \textbf{0.990} & 0.958 & \textbf{0.966} & 0.982 & \textbf{0.985} \\
EfficientNet-B0 & 0.989 & \textbf{0.992} & 0.983 & \textbf{0.986} & \textbf{0.982} & \textbf{0.982} & 0.984 & \textbf{0.985} & 0.955 & \textbf{0.958} & 0.978 & \textbf{0.981} \\ 
Xception & \textbf{0.976} & 0.975 & \textbf{0.979} & \textbf{0.979} & \textbf{0.976} & \textbf{0.976} & 0.977 & \textbf{0.978} & \textbf{0.939} & \textbf{0.939} & \textbf{0.969} & \textbf{0.969} \\ \hline
%
\end{tabular}
\caption{Performance in terms of accuracy on the test set for the different architectures with respect to the FF++ forgeries for RGB and RGB+GSD cases.}
\label{tab_rgb-feat}
\end{table*}

\paragraph{Training setting}
We implement SurFake in Pytorch. We model the deepfake detection as a binary classification problem and we train each classification network on an NVIDIA TITAN RTX. Specifically, we use a standard cross entropy loss with two classes, real and fake, for 30 epochs and batch size 32. We utilize SGD as optimizer with momentum $0.9$, weight decay $0.0001$ and learning rate $0.001$.

\subsection{Analysis of the proposed GSD feature performance}
\label{subsec_res_feature}
In this section, experimental results to evaluate the effectiveness of the proposed GSD feature will be presented. To better understand the capacity to provide distinctiveness between real and deepfake images, we have considered the activations obtained at the final layer of SurFake
, just before getting the output decision step. Therefore, we train MobileNetV2 to detect each face manipulation by using only our GSD feature as input without RGB. MobileNetV2 contains the initial fully convolution layer with 32 filters, followed by 19 residual bottleneck layers and ends with a linear layer with 1280 dimensional feature. We then plot these activations,
by resorting to T-SNE \cite{van2008visualizing}. As it can be seen in Figure \ref{fig:tsne_global} for all the 5 manipulation techniques of the FF++ dataset, it is possible to appreciate a certain separation between real samples (black dots) and fake ones (colored dots) which is coherent for all the different cases.
Even though our proposed GSD features between real and fake are often uniform (see Figure \ref{fig:face_uprightnet}), subtle differences can be perceived using a neural network, while, instead, being almost invisible to the human eye. The depicted T-SNE in Figure \ref{fig:tsne_global} clearly demonstrates how relevant these GSD patterns are (similar representations can be obtained with other network architectures). Although we are processing patches that describe surfaces of faces rather than canonical RGB face images, a significant amount of test samples have been correctly separated in the projection space. We also plot the ROC curves in the test set for all the face manipulations detected using MobileNetV2, in Figure \ref{fig:roc_global_mobilenetv2}. We deduce that in all the forgery techniques the AUC (Area Under Curve) is around 0.85, which is quite interesting considering the scarce visible information carried on this feature.
\begin{figure*}
\centering
    \begin{tabular}{ccc}
        \includegraphics[width=0.3\linewidth]{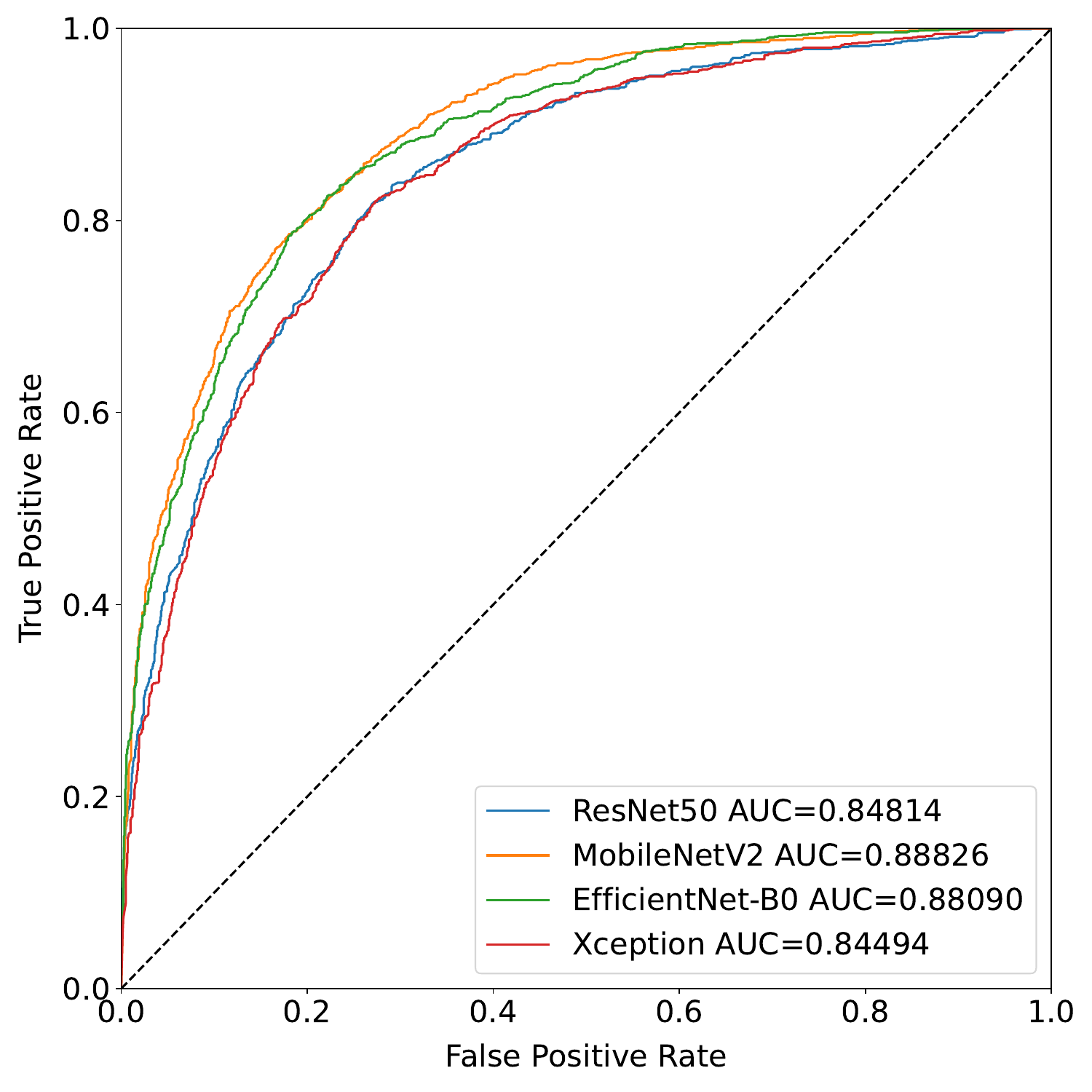} & \includegraphics[width=0.3\linewidth]{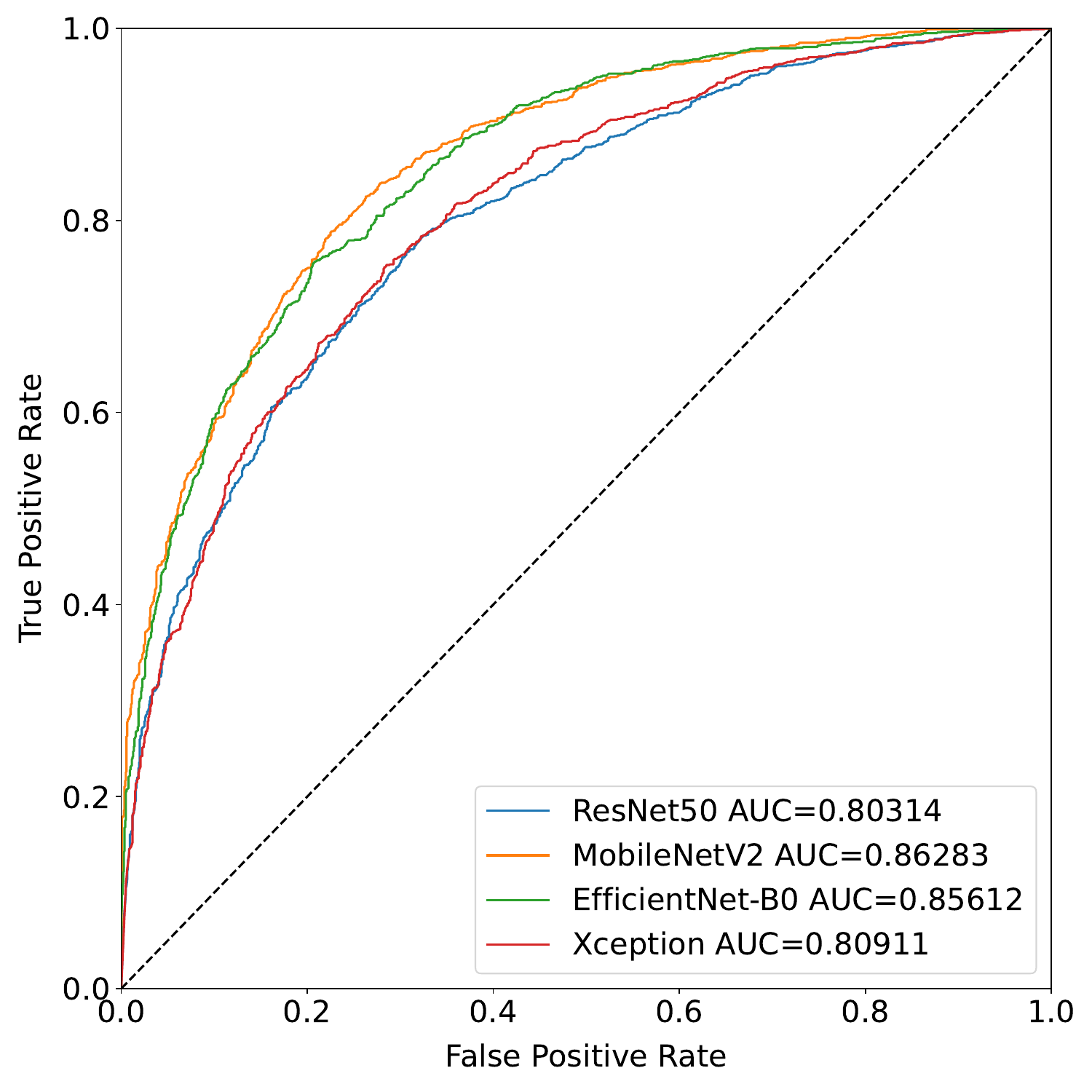} &  \includegraphics[width=0.3\linewidth]{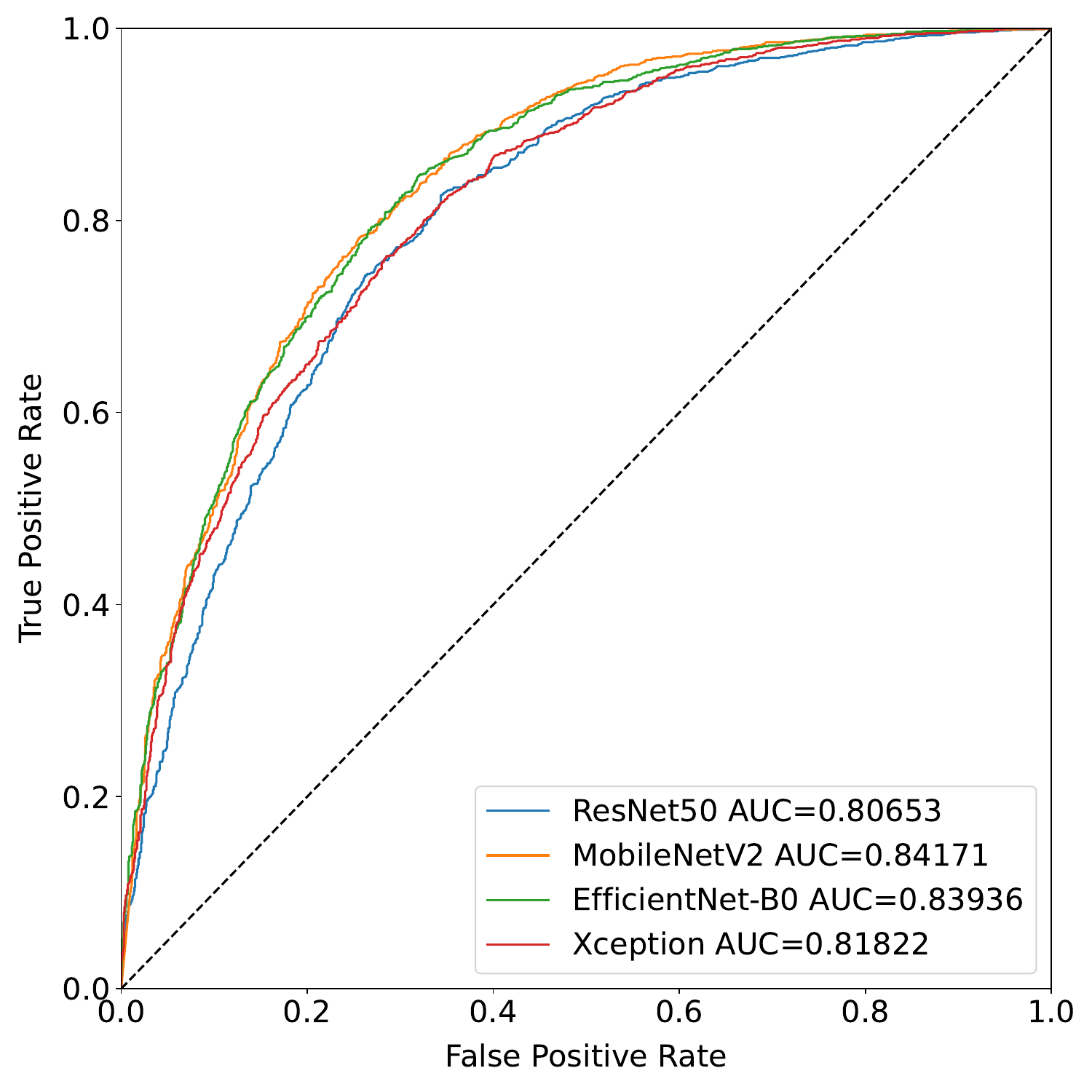} \\
         (a) DeepFakes (DF) & (b) Face2Face (F2F) & (c) FaceShifter (FSH) 
         \vspace{1em} \\
    \end{tabular}
    \begin{tabular}{cc}
         \includegraphics[width=0.3\linewidth]{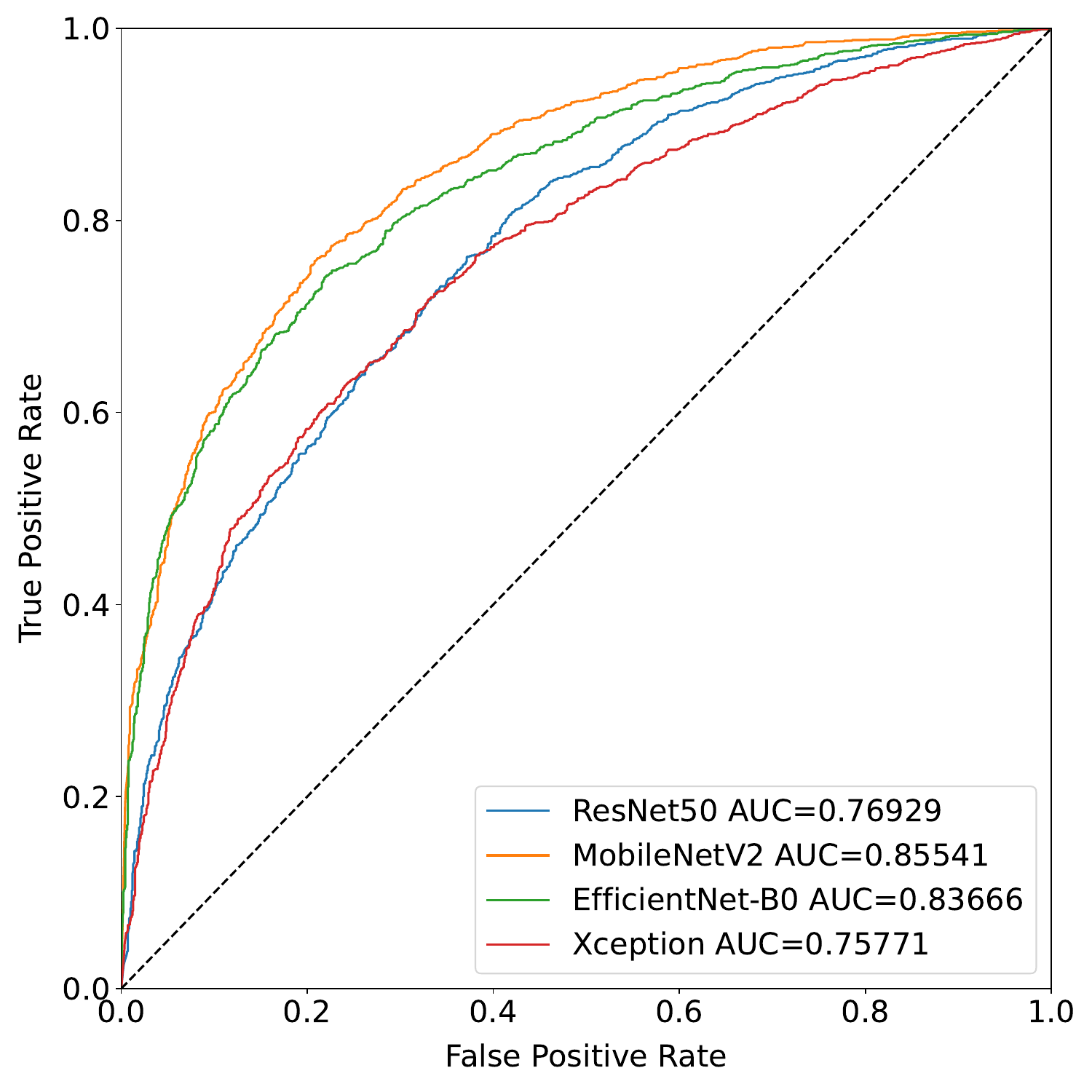} & \includegraphics[width=0.3\linewidth]{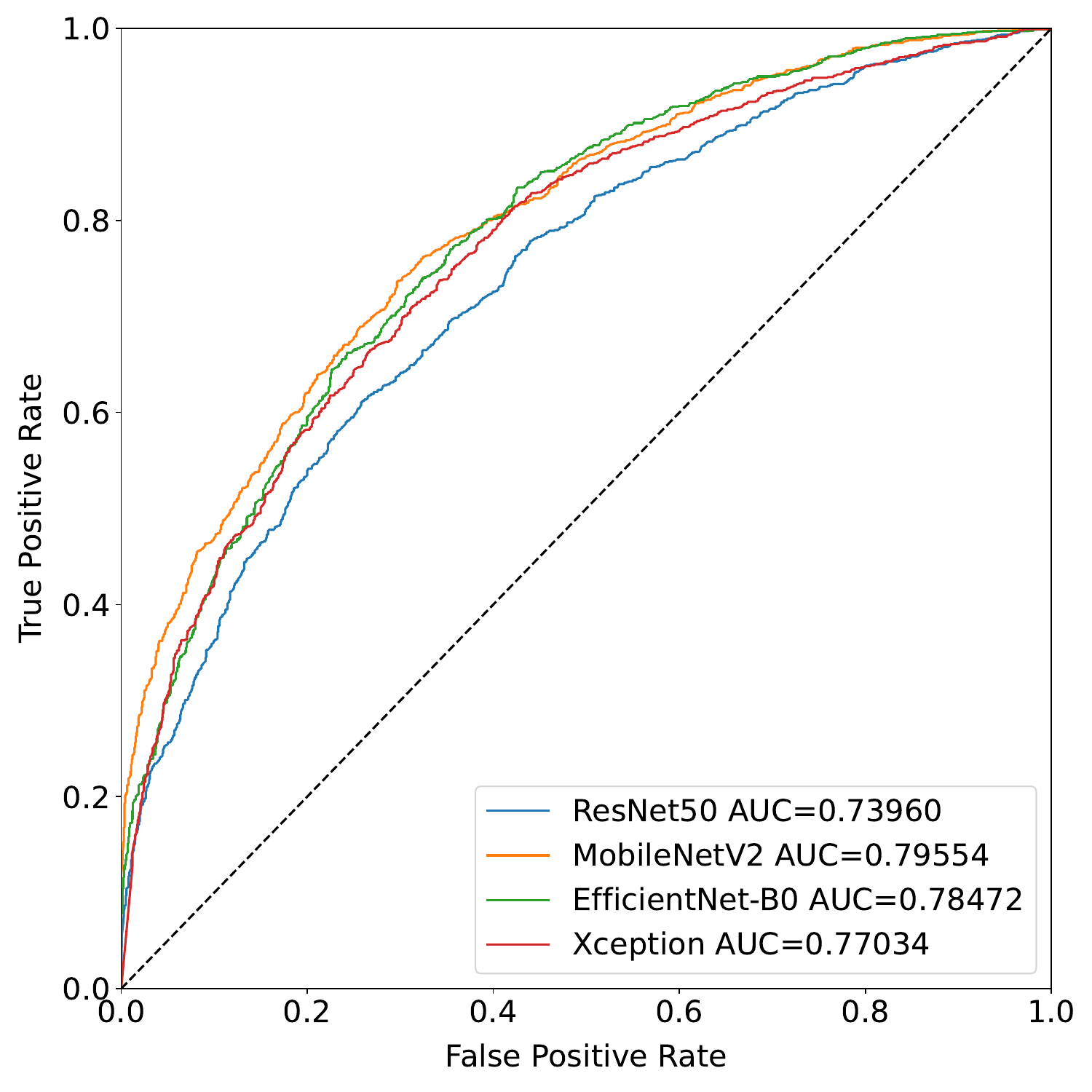} \\
         (d) FaceSwap (FS) & (e) NeuralTextures (NT) \\
    \end{tabular}
    \caption{Performance in terms of ROC curves on test set of the GSD features for each forgery in all the 4 architectures. We also report the Area Under Curve (AUC) in the legends.}\label{fig:roc_forgery}
\end{figure*}\\
Similarly, we have tried to make a quantitative evaluation of such a phenomenon and we have computed the accuracy values for all the five different distortions. To do that, we evaluate our approach, still using GSD features as unique input to the classification network, and we report our results for all the 4 architectures.
As listed in Table \ref{tab_featacc}, we can notice that in each case and, above all, coherently for different kinds of network architectures, an average accuracy around $0.75$ can be globally achieved.
We observe that GSD exhibits well distinctiveness in most of the manipulations for all the architectures with high accuracy, e.g. DeepFakes (DF), Face2Face (F2F) and FaceShifter (FSH) are well-detected. Performance on NeuralTextures (NT) are lower than the other manipulations and possibly this is due to the fact that NT deals with facial reenactment performed just around the mouth region \cite{rossler2019faceforensics++}.
Additionally, we report the ROC curves for each forgery of all the architectures in Figure \ref{fig:roc_forgery}, which highlights the efficacy of the GSD features in most cases, as the average Area Under Curve is above $0.80$, except for NT which gets an average of $0.77$.
We can also observe that EfficientNet-B0 and MobileNetV2 report higher AUC in most of the forgeries. 

\subsection{Composing GSD with RGB frames}
\label{subsec_res}
Hereafter, we will present the results obtained by composing the 3-channel GSD with the RGB frames that are usually adopted as primary source of information in most deepfake detection methods. This has been done in order to understand if the proposed GSD feature is able to provide an improvement in deepfake detection, thanks to the fact that it takes into account geometrical components related to the acquisition scene. In this case the diverse network architectures have been trained by receiving as input a 6-channel tensor composed of 3 RGB bands concatenated with the 3 GSD channels. The achieved performances in terms of detection accuracy are listed in Table \ref{tab_rgb-feat}. As it can be seen, by looking at the last column of the table, a general increment is registered on average. Due to the fact that accuracy values are already quite high such improvement is rather limited but, what is interesting is that it is consistent for all the five forgeries and coherent for all the different network architectures that we considered. In particular, if we look at the NT case that usually appears to be more difficult to treat, it is possible to appreciate that an overall trend of increment is achieved for all the four networks. It is worth to point out that for the Xception model a substantial similar behavior is registered and the GSD feature does not seem to bring a relevant advancement.
Since Xception gets input images of $299 \times 299$ but our original patches are of $224 \times 224$, both RGB and GSD have to be upscaled and fit them to the bigger resolution. That potentially adds interpolation artifacts.
Indeed, ResNet50, that can directly get as inputs the original patches (as well as the other selected architectures), even though it has a different architecture from Xception but with comparable number of trainable parameters and performance reported in ImageNet, obtains an overall improvement when RGB is concatenated along with GSD. EfficientNet-B0, still with a different architecture but with similar performance and with one fifth of the trainable parameters than ResNet50 and Xception, got more improvement, with an average accuracy across all forgery manipulations of 0.981 (i.e. $+0.3\%$ using the percentage notation).
Overall, we notice that our approach employing either EfficientNet-B0 or MobileNetV2 gets more performance improvement. This is also confirmed, by looking at the ROC curves in Figure \ref{fig:roc_forgery}, where our approach is only trained with the GSD features. These two architectures show a superior trend of the ROC curves with respect to others, and with the corresponding Area Under Curve values highest among the manipulations. 
Such behavior demonstrates that our proposed GSD feature introduced in a classification network can benefit the detection of deepfakes.

\section{Conclusions}
\label{sec:conclusions}
In this paper we proposed a novel deepfake approach named SurFake able to detect face manipulations at frame level. To do that, we introduced the use of Global Surface Descriptor (GSD) as feature that accounts for the camera acquisition process which marks permanently the image. In particular, we exploited the characteristics of the surfaces in which pixels belonging to horizontal or vertical areas of the image have a proper direction and intensity with respect to a global coordinate system. 
We tested SurFake with 4 different architectures on FaceForensics++, which contains 5 different face manipulations. We demonstrated that our proposed GSD features alone allow a classifier to reach around 75\% of accuracy on average; furthermore, we tested our proposed pipeline by using RGB frames and GSD together as input and we got an overall improvement, though limited, for all the diverse face manipulations.

As future works, we consider to deal with larger cropped patches in order to possibly improve the effectiveness of GSD and also to make some data augmentations, e.g. random crop. We will investigate other geometric information estimated by UpRightNet methodology, e.g. the local surface geometry which is directly tied to the local coordinate system. Finally, we will carry out further experiments on other deepfake datasets.

\paragraph{Acknowledgements}
This work was partially supported by the H2020 project AI4Media (GA n. 951911) and by the project SERICS (PE00000014) under the NRRP MUR program funded by the EU - NGEU.

\clearpage
\bibliographystyle{ieee_fullname}
\bibliography{egbib}
\balance

\end{document}